# Communicative Agents for Slideshow Storytelling Video Generation based on LLMs

Jingxing Fan[a,*], Jinrong Shen[a], Yusheng Yao[a], Shuangqing Wang[b], Qian Wang[b], Yuling Wang[b]

[a] *State Key Laboratory of Integrated Chips and Systems, College of Integrated Circuits and Micro-Nano Electronics, Fudan University*
[b] *Shanghai Turtle Technology Co., 69 Yuan Feng Road, Baoshan District, Shanghai 200433, China*

There are no conflicts of interest regarding the publication of this paper. This work was conducted independently and was not supported by any external funding or grants. There are no financial or personal relationships with other people or organizations that could inappropriately influence or bias the content of the paper.
*Corresponding author: Jingxing Fan.*
*Jingxing Fan* is with State Key Laboratory of Integrated Chips and Systems, College of Integrated Circuits and Micro-Nano Electronics, Fudan University, 220 Handan Road, Yangpu District, Shanghai, 200433, China. (e-mail: fanjx23@m.fudan.edu.cn).

*Abstract—*
With the rapid advancement of artificial intelligence (AI), the proliferation of AI-generated content (AIGC) tasks has significantly accelerated developments in text-to-video generation. As a result, the field of video production is undergoing a transformative shift. However, conventional text-to-video models are typically constrained by high computational costs.

In this study, we propose Video-Generation-Team (VGTeam), a novel slide show video generation system designed to redefine the video creation pipeline through the integration of large language models (LLMs). VGTeam is composed of a suite of communicative agents, each responsible for a distinct aspect of video generation, such as scriptwriting, scene creation, and audio design. These agents operate collaboratively within a chat tower workflow, transforming user-provided textual prompts into coherent, slide-style narrative videos.

By emulating the sequential stages of traditional video production, VGTeam achieves remarkable improvements in both efficiency and scalability, while substantially reducing computational overhead. On average, the system generates videos at a cost of only $0.103, with a successful generation rate of 98.4%. Importantly, this framework maintains a high degree of creative fidelity and customization.

The implications of VGTeam are far-reaching. It democratizes video production by enabling broader access to high-quality content creation without the need for extensive resources. Furthermore, it highlights the transformative potential of language models in creative domains and positions VGTeam as a pioneering system for next-generation content creation.

*Index Terms—*AI agents, Human-AI interaction, Large Language Models, Video Generation

## I. INTRODUCTION

THE evolution of video production is at a pivotal crossroads driven by artificial intelligence and its expanding capabilities. Historically, the creation of video content has been both resource-intensive and cost-prohibitive, and accessible mainly to those with substantial budgets and technical expertise. As a result, the ability to craft and disseminate video narratives has historically been limited to a relatively small cohort of creators and organizations, often constrained by the challenges of budgeting, scheduling, equipment availability, staffing, and the complexities of outsourcing production efforts [1].

Concurrently, the advent of artificial intelligence has ushered video production into a new era, wherein text-to-video models have introduced an innovative paradigm for content creation—one that is inherently more accessible to a broader range of creators [2–7]. These models significantly streamline the video generation process and reduce reliance on extensive technical expertise. Nevertheless, they are accompanied by a number of critical limitations. Chief among these are the substantial computational costs, inconsistent video quality, and limited capacity for effective human intervention during generation. Moreover, the development and deployment of such models are heavily contingent upon large-scale, high-quality datasets, the construction of which is both resource-intensive and restricted in availability. These challenges collectively underscore the need for continuous advancement in text-to-video technologies in order to fully realize their narrative and creative potential.

The emergence of large language models (LLMs) has further accelerated the development of text-to-video generation algorithms. Recent studies have demonstrated the use of LLMs—such as ChatGPT [8]—in roles akin to video directors [9] and dynamic scene managers [10]. Despite these promising applications, the integration of LLMs into video production workflows remains in its nascent stages, with much of their potential yet to be fully explored. Beyond their capabilities in text or feature generation, LLMs can also function as communicative agents [11], enabling more flexible and interactive systems. Recent research has validated the feasibility of constructing systems composed entirely of LLMs [12–16]. Moreover, simulations of human societies [17,18] and virtual production companies [19,20] powered by LLMs are increasingly becoming viable, further highlighting their transformative potential in both creative and organizational contexts.

To address the prevailing challenges in video production and to harness the full potential of large language models (LLMs), we propose Video-Generation-Team (VGTeam), a multi-agent system designed for fully automated slideshow storytelling video generation. This system leverages a set of communicative agents that collaboratively manage different aspects of the video creation pipeline. By relying solely on Application programming interface-based (API) operations, VGTeam transforms simple textual input into complete video output in a cost-effective and time-efficient manner, without the need for dedicated models or intensive computational resources.

In summary, the primary contributions of this work are as follows:

- We present a slideshow storytelling video production system in which distinct AI agents perform specialized roles, such as directing, editing, and visualizing, driven entirely by large language models and API interactions.
- Our framework utilizes APIs for image generation, voice synthesis, and music creation, which facilitate an automated production pipeline without the need for conventional computational models.
- We conducted 300 independent video generation experiments to evaluate the robustness and adaptability of VGTeam. By varying user input across a wide spectrum of prompt lengths and styles, we systematically investigated the impact of input diversity on system performance, as well as the behavioral differences among multiple large language models integrated within the framework.

Fig 1 illustrates the overall data flow within the framework. The system begins by receiving a user-provided textual prompt, which subsequently triggers a set of AI agents responsible for generating distinct video components. Finally, the individual video components are integrated to produce a slideshow storytelling video.

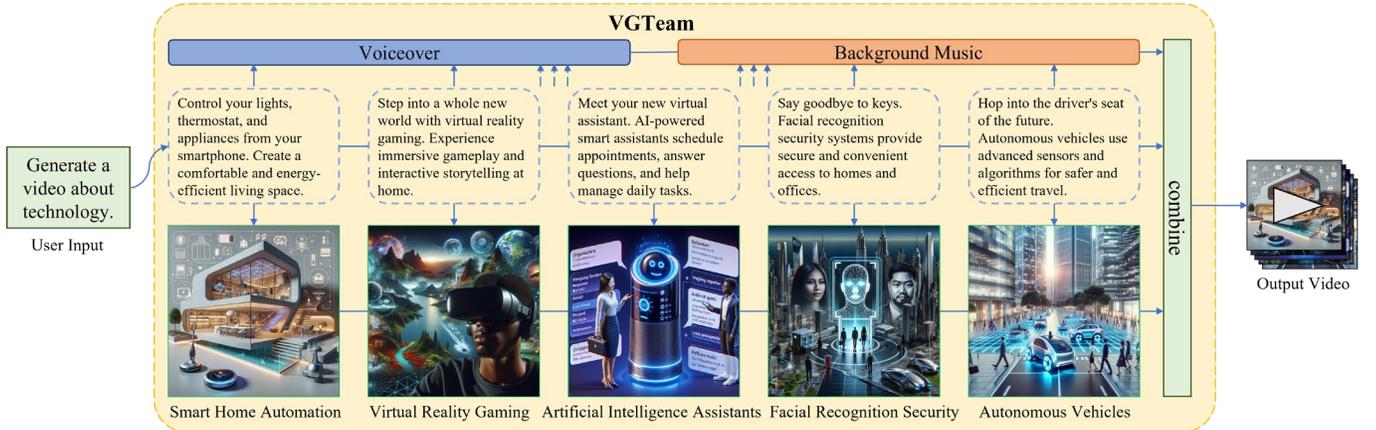

**Fig. 1.** Data flow of VGTeam system. The workflow begins with the reception of a textual input from the user. Subsequently, individual AI agents within the system generate video subtitles, images, and background music through dedicated API calls. Finally, these components are integrated to produce the final video output.

## II. VGTeam

In the domain of video production, conventional text-to-video methods have laid the groundwork for automated content creation. However, these methods often encounter challenges related to limited adaptability across tasks and suboptimal production efficiency. The VGTeam framework addresses these challenges by drawing inspiration from LLM-based virtual communication systems, wherein AI agents are assigned to specific roles within the production pipeline. This role-based delegation enhances operational efficiency and reduces production costs, thereby overcoming the constraints of traditional methods.

By explicitly assigning distinct responsibilities—such as director, editor, painter, and composer—VGTeam mitigates common issues associated with LLMs in complex workflows, including ambiguous instructions and overly long task sequences. Furthermore, the system leverages API-based calls to generate video components, significantly lowering the computational demands associated with video generation (Fig 2). This API-centric strategy not only broadens the applicability of LLMs within creative domains but also demonstrates the system's strong potential for scalability and adaptability in content creation.

### A. Chat Tower

To address the challenges posed by linguistic ambiguity and overly broad instructions during agent interactions—which can lead to execution errors or hallucinations [21]—VGTeam adopts a novel agent interaction structure inspired by the waterfall model commonly used in software engineering [19], as illustrated in Fig 2.

During the conceptualization phase of video content, we introduce the Chat Tower architecture, wherein agents engage in a sequential and structured dialogue to produce all necessary textual elements for video generation, including video captions, image prompts, and music prompts. In the subsequent creation phase, the system invokes a series of text-to-X APIs to obtain the visual and auditory components required for the final output.

The Chat Tower serves as the central mechanism governing the logic of agent communication. In this architecture, the user interacts solely with the agent director, ensuring thematic consistency and narrative coherence. The director, in turn, formulates task-specific objectives and directives for each downstream agent, guiding them to contribute toward a unified video concept. The subsequent agents—editor, painter, and composer—carry out their designated responsibilities in coordination with one another to ensure the overall integrity of the production. Notably, the video caption generated by the agent editor is shared with both the agent painter and agent composer to maintain stylistic alignment between the visual, auditory, and narrative elements of the video.

Furthermore, all interactions within the Chat Tower are transparently logged, thereby making the decision-making processes among agents traceable and enabling real-time error diagnosis. This logging mechanism facilitates human oversight and intervention when necessary, ensuring that each output adheres to quality and content requirements.

In summary, the Chat Tower represents a structured yet flexible framework for multi-agent communication and task-oriented collaboration. It fosters clear inter-agent coordination, supports seamless workflow execution, and significantly enhances both the quality and efficiency of slide show video generation.

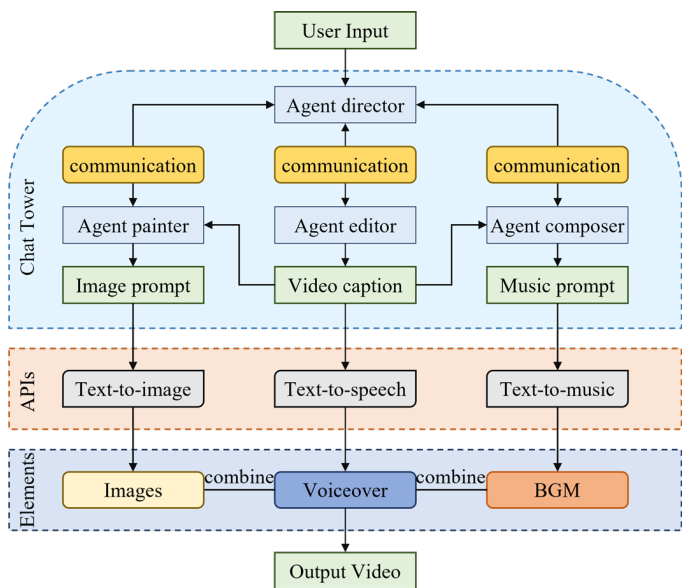

**Fig. 2.** System architecture of VGTeam. The process begins with the agent director receiving user input. The director then coordinates with downstream agents—the editor, painter, and composer—each responsible for generating textual descriptions corresponding to specific components of the video. These textual outputs are subsequently used to invoke appropriate APIs for the generation of visual and auditory elements, which are finally integrated into a complete video output.

### B. Detailed Realization

#### a) Role Specialization

In the VGTeam framework, the transformation of a general-purpose large language model (LLM) into a role-specific agent constitutes a critical component of the video production pipeline. This transformation is achieved through a process known as prompt engineering, which plays a key role in adapting the broad generative capacity of LLMs to the specific and context-sensitive demands of video production tasks.

As shown in Fig 3, role specialization within VGTeam is achieved through the systematic design of prompts, which function as explicit instructions guiding the behavior of LLMs in accordance with assigned roles. These prompts are engineered to define the scope and objectives of the task, expected inputs and outputs, along with standards for successful completion. This prompt engineering is essential to transform the general capabilities of LLMs into specialized agents that fulfill the requirements of a defined role within the video production workflow.

The process of defining roles through prompts was meticulously tailored to each project. The prompts include:

Task Objectives: Clearly defined goals that an agent must achieve, such as developing a storyline or editing a scene to match a certain tempo.

Input and Output Requirements: Specific instructions on the information the agent receives and the expected format of their deliverables.

Performance Standards: Benchmarks that determine the quality and relevance of the output, ensuring that it meets the standards.

By applying these prompts, VGTeam ensures that each agent operates within a clearly defined role. This innovative approach not only harnesses the full potential of LLMs but also significantly reduces the risk of content misalignment or production errors.

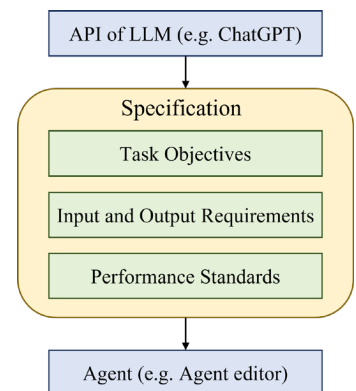

**Fig. 3**. Role specialization in VGTeam. The definition of agent roles begins with the invocation of the LLM API. By providing a constructed system prompt that outlines task requirements and contextual information, each agent is assigned a distinct functional role within the video production pipeline.

#### b) Memory Stream for Continuity

To ensure coherent context and unified objectives throughout agent interactions—while mitigating the common issue of forgetting in large language models (LLMs)—we introduce a key subsystem termed the memory stream. This component is designed to capture and persist the outcomes of prior dialogues, thereby offering agents a reliable reference framework that supports informed decision-making during subsequent tasks.

For instance, such continuity is particularly crucial for the agent editor (Fig 4). During scriptwriting, the memory stream retains an accessible record of prior instructions, including narrative objectives, tone, and pacing requirements. Referencing this repository allows the editor to maintain consistency with the intended narrative structure and the director's vision.

The persistence of the memory stream also mitigates potential deviations that may arise when agents operate independently without full awareness of the overall process. By enabling agents to recall and reference the outputs of their counterparts, the framework fosters a more cohesive and coordinated workflow.

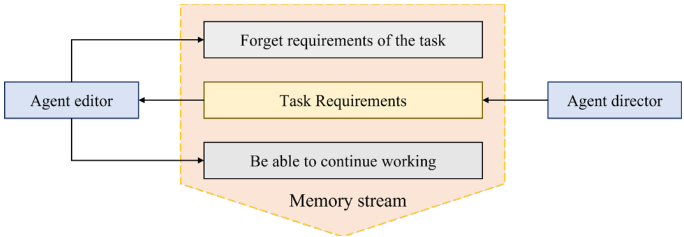

**Fig. 4**. Memory Stream in VGTeam. The memory stream preserves the director's instructions, allowing the editor to recall specific guidance and ensure that content creation remains consistent with the project's original vision and objectives.

c)   Iterative Approval for Quality Assurance

Within the Chat Tower in VGTeam, the agent director plays a crucial role in the quality gatekeeper, ensuring that each output from the agents aligns with the project's quality standards. This section describes the iterative approval mechanism that is fundamental to VGTeam's commitment to maintaining high standards of quality.

During agent interactions within the system, the director reviews the output from the editor, painter, and composer agents. If the output does not meet the predefined criteria, the director provides specific feedback, which is then used by agents to refine their work. This iterative cycle continues until the director deems the output satisfactory and grants approval.

For instance, as shown in Fig. 5, the editor may submit a draft of the video script, which the director reviews against the project's narrative goals. If the script requires refinement to better align with the intended narrative, the director provides revision notes. The editor then revises the script based on this feedback to ensure coherence with the overarching storyline. This process is repeated as necessary, with the Memory Stream serving as a reference to maintain continuity with the previous versions and feedback.

The effectiveness of this process is attributed to the director's ability to provide clear and actionable feedback, enabling agents to iteratively refine their outputs through collaborative improvement. This structured yet flexible approach enhances the overall quality of the generated video content while promoting adherence to broader social and ethical constraints. It underscores the central role of the director within the Chat Tower framework—not only as the initiator of tasks but also as the arbiter of quality—who guides the project toward successful completion.

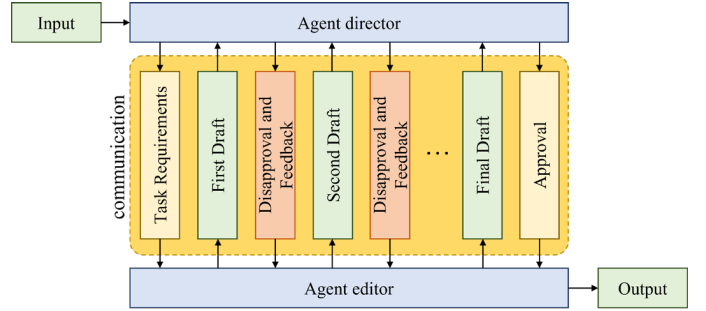

**Fig. 5**. Iterative Approval Process. The director will provide feedback, for example prompting the editor to revise the narration. This cycle continues until the director is satisfied with the modifications and gives the final approval.

d)   API-Driven Video Generation system

VGTeam leverages API-based services to construct the individual components of the video. This system allows the framework to interface with external APIs for image generation, voice synthesis, and music composition, thereby eliminating the need for computationally intensive models. Nevertheless, the system's dependence on third-party APIs may lead to fluctuations in video quality and production performance.

## III. EXPERIMENTS

In our experiments, we employed DeepSeek-V3, ERNIE 4.5-Turbo, and Qwen3-235B as the primary APIs for role definition and script interaction. For multimedia content generation, ERNIE-iRAG was utilized for image synthesis, Baidu's TEXT2AUDIO API for voiceover production, and Manolis Teletos's TEXT-to-MUSIC API for background music generation. The experimental environment was based on Python 3.11, with MoviePy 1.0.3 used for video editing and post-processing. All experiments were conducted on a system equipped with an NVIDIA GTX 1650 GPU, ensuring consistent performance and reliability.

*A.  Case Study*

In this section, we evaluate the stability and output efficacy of VGTeam. To demonstrate the system's adaptability across different language models, we conducted three sets of experiments, each consisting of 100 identical input prompts. These 100 prompts were selected from the most popular video topics in the YouTube-8M dataset [22, 23], and were evenly distributed across the following categories: Class 1 – Vehicle, Class 2 – Concert, Class 3 – Association Football, Class 4 – Animal, and Class 5 – Food. To assess the impact of input length on performance, we defined two input types: short prompts (1–5 words) and long prompts (more than 10 words). The experimental results are summarized in Fig. 6.

Fig. 6 illustrates the statistical distribution of the 300 experimental runs. The overall failure rate was 1.7%, with failures primarily attributed to three factors: network instability, character confusion states, and infinite loops. Character confusion refers to rare instances in which an AI agent misinterprets the task—such as assuming it cannot process non-textual elements—and can typically be mitigated through prompt optimization. Infinite loops tend to arise when the "director" role fails to generate a satisfactory outcome, triggering repeated attempts. While network instability is challenging to control without high-quality internet infrastructure, both character-related errors and looping issues can be reduced through effective prompt engineering and loop-limiting strategies.

Notably, all observed failures occurred in trials with short prompt inputs, indicating that more detailed prompts (i.e., longer inputs specifying desired video content) enhance the stability of the generation process. Additionally, inappropriate content appeared in 22.7% of the outputs. Such content typically includes repetitive visual elements [24] or semantically incoherent image prompts generated by the "painter" role. Although inappropriate content is often identifiable by built-in content moderation mechanisms in the APIs, it can further be minimized through stricter constraints on the painter module.

Overall, the successful generation rate was high at 98.4%, with 75.7% of the videos being properly generated, indicating that the system functions effectively under most conditions.

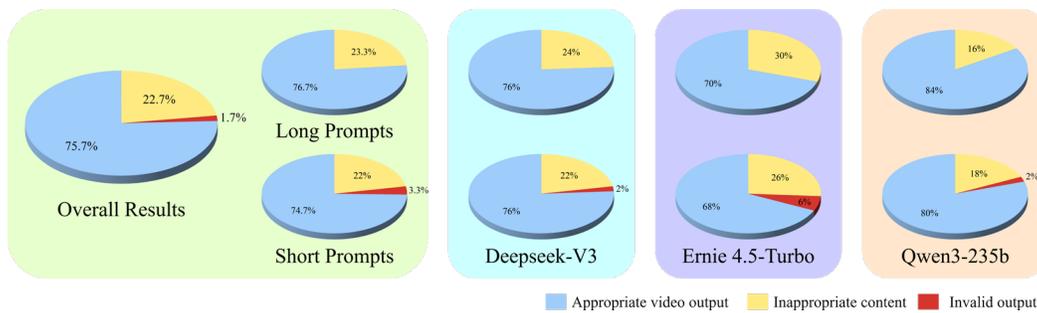

**Fig. 6.** Outcome Distribution of VGTeam. This figure presents the outcome distribution of VGTeam across 300 trials, classified into three categories: appropriate outputs, inappropriate content, and invalid outputs. The results are further disaggregated by language model and input length, distinguishing between short prompts and long prompts.

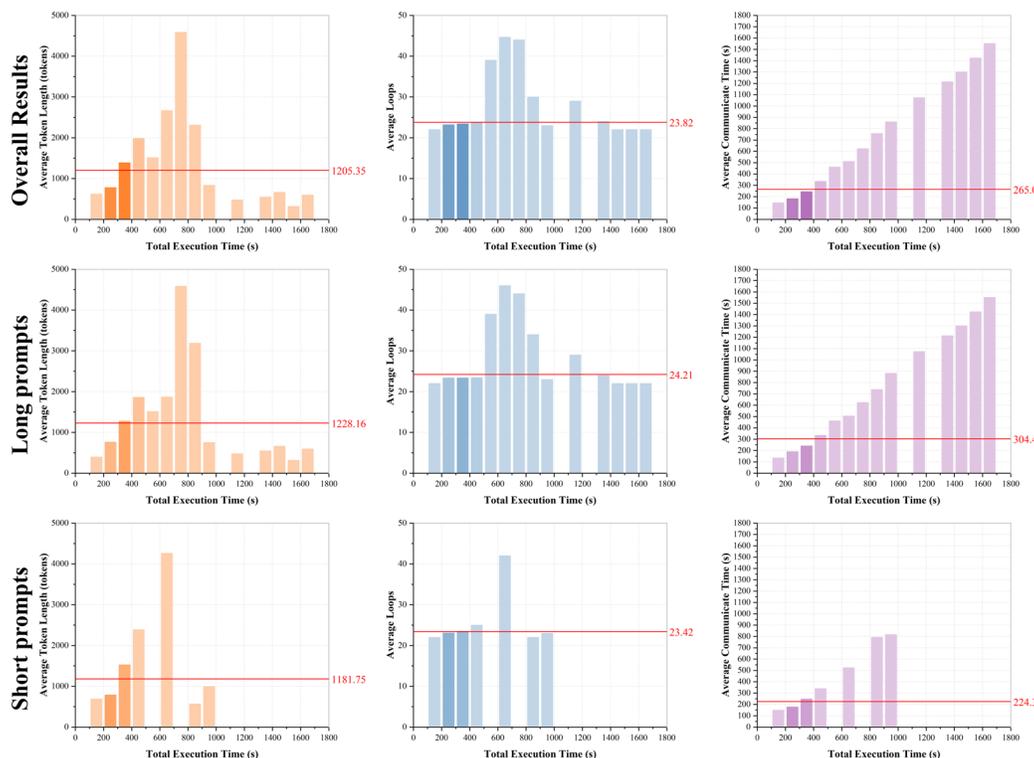

**Fig. 7.** Temporal Distribution of Key Metrics Across Prompt Types. Average token length, loop count, and communication time under different prompt types (overall, long, and short) across total execution time intervals. Red lines indicate metric averages. Long prompts show greater variability, while short prompts yield more stable runtimes.

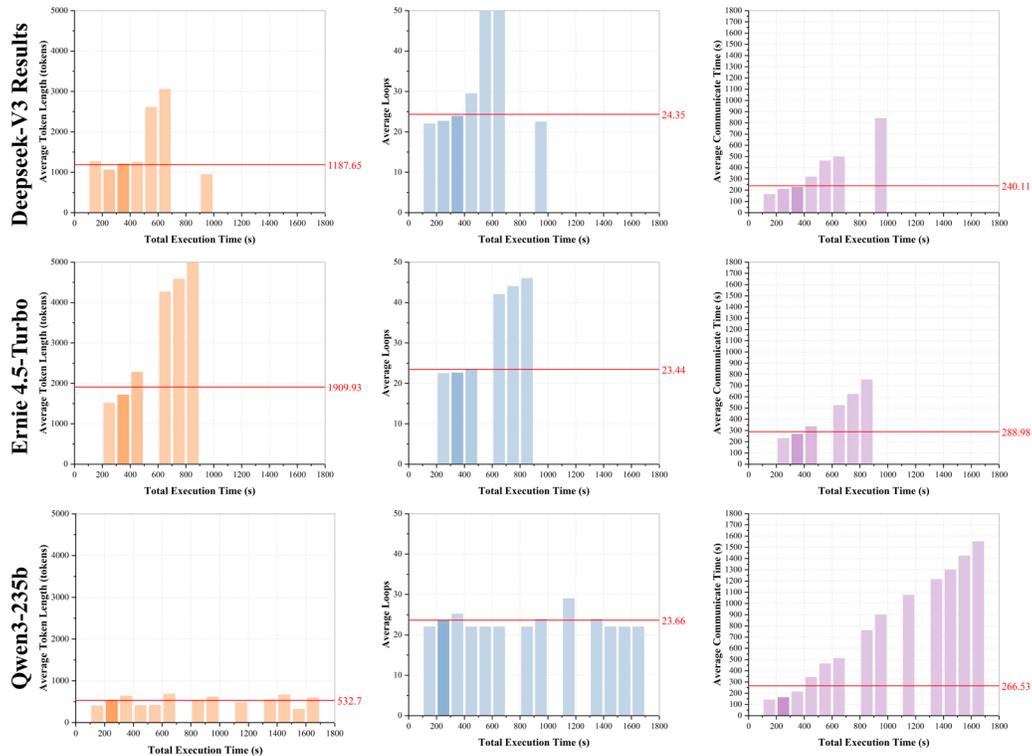

**Fig. 8**. Performance Comparison Across Language Models. Comparison of average token length, loop count, and communication time across different language models. Deepseek-V3 and Ernie 4.5-Turbo demonstrate consistent behavior; Qwen2-35B shows lower token length but broader execution time distribution.

*B. Duration Analysis*

The duration analysis is founded on a comprehensive evaluation of 300 experimental runs, detailed through three histograms (Fig. 7-Fig. 8) each representing a distinct operational metric: "token length", "loops", and "communicate time". The x-axis on these histograms signifies the total runtime segmented into intervals, where the depth of color correlates with the number of cases within each interval.

Fig. 7 presents the temporal distribution analysis across 300 experimental runs, and the results reveal a clear positive correlation between total execution time and communication time, with an overall average communication duration of 265.04 seconds. This trend suggests that prolonged runtimes are primarily influenced by extended API interactions and message exchanges among agents. In contrast, the trends for token length and loop count are more nuanced. Besides, longer prompts tend to result in slightly higher token lengths (average: 1228.16) and loop counts (24.21). Comparatively, short prompts yield slightly lower average values—token length of 1181.75 and loop count of 23.42—but demonstrate a more centralized and stable distribution of total execution times. This consistency is further reflected in the communication time for short prompts, averaging 224.3 seconds, compared to 304.42 seconds for long prompts.

These findings suggest that short prompt inputs, although less semantically rich, tend to elicit more predictable and efficient system behavior by reducing communication overhead and loop frequency. However, this simplicity may also increase the likelihood of output failure due to insufficient contextual information provided to the system. Conversely, long prompts, while increasing the likelihood of generating high-quality and contextually complete outputs, also introduce variability in execution, particularly in communication overhead and loop iterations.

Fig. 8 illustrates the performance breakdown across three language models—Deepseek-V3, Ernie 4.5-Turbo, and Qwen3-235b. This model-specific analysis highlights distinct behavioral patterns in how each model processes video generation tasks.

Ernie 4.5-Turbo exhibits the highest average token length (1909.93 tokens), reflecting its tendency to generate more verbose outputs. It also demonstrates a concentrated execution time distribution centered around 200–400 seconds, accompanied by moderate communication latency (average: 288.98 seconds). Deepseek-V3 shows slightly lower token lengths (average: 1187.65) but maintains a comparable loop count (24.35), indicating balanced but iterative task resolution behavior. In contrast, Qwen3-235b displays a markedly different pattern: its average token length is substantially lower (532.7), yet its execution time is more widely distributed, often extending beyond 1200 seconds. This suggests that although Qwen3-235b produces shorter outputs, it is more prone to prolonged internal processing or inefficient communication cycles (communication time average: 266.53 seconds).Overall, the differences among models suggest trade-offs between verbosity, execution efficiency, and system stability. Models like Ernie 4.5-Turbo and Deepseek-V3 may be more suitable for tasks requiring detailed narrative generation, while Qwen3-235b might prioritize conciseness at the cost of longer runtime dispersion. These findings underscore the importance of model selection in optimizing both performance and user experience within the VGTeam pipeline.

It is noteworthy that during 2025, specifically from March to May, the average cost of generating each video was approximately $0.103. This figure was calculated by dividing the total cost incurred from API calls during that period by the number of videos generated.

To provide a better representation of the user inputs and associated parameters utilized in our experiments, Tab.1 selects 10 samples, consisting of 5 long inputs and 5 short inputs.

**Tab. 1**. Video Generation Case Evaluation

| User Input | Total Loops | Total Token Length | Communicate Time(s) | Total Time(s) |
|---|---|---|---|---|
| This owl's head-turning trick is both creepy and amazing—here's how it works | 22 | 419 | 169.42 | 275.97 |
| How dolphins communicate using clicks, whistles, and underwater body language | 24 | 547 | 1214.34 | 1364.73 |
| Following a baby elephant's journey from rescue to rewilding | 32 | 668 | 170.72 | 279.82 |
| What your dog's tail movements actually mean according to science | 44 | 1078 | 203.11 | 313.98 |
| The incredible migration of monarch butterflies through three countries and generations | 22 | 440 | 182.38 | 287.07 |
| One Paw at a Time | 26 | 502 | 159.94 | 255.80 |
| Silent Hunters of Night | 22 | 571 | 163.03 | 283.93 |
| In the Lion's Shadow | 22 | 585 | 163.70 | 299.47 |
| Penguins in a Desert | 24 | 543 | 157.18 | 260.35 |
| Why Do Cats Stare | 22 | 361 | 136.61 | 239.50 |

## IV. Discussion

The VGTeam system represents a notable advancement in the domain of automated video generation, offering a highly efficient and cost-effective alternative to traditional methods. Nevertheless, this emerging framework also presents several limitations that warrant further investigation and refinement.

One primary concern lies in VGTeam's dependence on large language models (LLMs), which can introduce a degree of unpredictability into the content generation process. This stochastic behavior may result in output inconsistencies, whereby similar inputs do not reliably produce similar outputs. Our experiments further reveal that the type of LLM integrated into the system significantly affects key performance metrics, including time consumption, token length usage, and the overall stability of video generation. Future iterations could mitigate these issues by incorporating more deterministic control mechanisms or by fine-tuning prompt engineering strategies to enhance output consistency and system reliability across different LLM configurations.

Additionally, VGTeam currently relies on static imagery supplemented with basic animation techniques. While this approach streamlines production, it constrains the expressive capacity and visual dynamism of the generated videos. Future enhancements might explore the integration of more sophisticated visual technologies—such as 3D modeling, keyframe animation, or motion capture—to produce more immersive and engaging visual narratives.

Finally, as VGTeam matures, it becomes increasingly important to address the legal and ethical considerations surrounding automated video production. Potential issues include copyright violations, misrepresentation, and the unethical manipulation of visual or narrative content. Establishing clear guidelines, safeguards, and oversight mechanisms will be essential to ensure that such technologies are applied responsibly and align with societal standards.

## V. Conclusion

In this study, we introduced VGTeam, an AI agent-based framework for slideshow storytelling video generation. VGTeam makes three core contributions to LLM-based video generation. First, it introduces a multi-agent system where each agent takes on a clear role, such as directing, editing, or creating visuals, and works through natural language and API calls. Second, it uses an API-driven workflow that supports automated and efficient video production without requiring heavy computational resources. Besides, the inclusion of a human-in-the-loop mechanism allows for flexible intervention, helping ensure that outputs align with user intent and meet quality standards. Finally, through experiments with varied user inputs, we observed that prompt diversity and the choice of LLM significantly influence time consumption, token usage, and generation stability. Together, these features make VGTeam an effective and adaptable framework for AI-assisted video creation, achieving a 98.4% success rate at an average cost of just $0.103 per video.

Moving forward, we will focus on refining the system stability, enhancing the complexity of the generated content, and addressing ethical considerations. VGTeam represents a step forward in AI-driven multimedia creation, combining language understanding with visual storytelling. It offers a foundation for automating tasks such as text-to-video generation, enabling efficient production of slideshow videos and other creative content. This work opens up promising directions for future research in scalable and accessible video generation systems.

DATA AVAILABILITY

The videos generated in this work are available through: https://github.com/hmfjx123/VGTeam.

Our data are available upon reasonable request.